v1. Aug. 22, 2021

# Explainable Machine Learning using Real, Synthetic and Augmented Fire Tests to Predict Fire Resistance and Spalling of RC Columns

M.Z. Naser[1,*], V.K. Kodur[2]

[1,*]Assistant Professor, Glenn Department of Civil Engineering, Clemson University, Clemson, SC, USA.
AI Research Institute for Science and Engineering (AIRISE), Clemson University, Clemson, SC 29634, USA
E-mail: mznaser@clemson.edu, m@mznaser.com, Website: www.mznaser.com

[2]University Distinguished Professor, Department of Civil and Environmental Engineering, Michigan State University, East Lansing, MI, Email: kodur@egr.msu.edu, Website: https://vkodur.wixsite.com/kodur

**ABSTRACT**
This paper presents the development of systematic machine learning (ML) approach to enable explainable and rapid assessment of fire resistance and fire-induced spalling of reinforced concrete (RC) columns. The developed approach comprises of an ensemble of three novel ML algorithms namely; random forest (RF), extreme gradient boosted trees (ExGBT), and deep learning (DL). These algorithms are trained to account for a wide collection of geometric characteristics and material properties, as well as loading conditions to examine fire performance of normal and high strength RC columns by analyzing a comprehensive database of fire tests comprising of over 494 observations. The developed ensemble is also capable of presenting quantifiable insights to ML predictions; thus, breaking free from the notion of "black-box" ML and establishing a solid step towards transparent and explainable ML. Most importantly, this work tackles the scarcity of available fire tests by proposing new techniques to leverage the use of real, synthetic and augmented fire test observations. The developed ML ensemble has been calibrated and validated for standard and design fire exposures and for one-, two-, three- and four-sided fire exposures thus; covering a wide range of practical scenarios present during fire incidents. When fully deployed, the developed ensemble can analyze over 5,000 RC columns in under 60 seconds; thus, providing an attractive solution for researchers and practitioners. The presented approach can also be easily extended for evaluating fire resistance and spalling of other structural members and under varying fire scenarios and loading conditions and hence paves the way to modernize the state of this research area and practice.

## 1.0 INTRODUCTION

Exposure to elevated temperatures, as encountered in fire, adversely affect the properties of construction materials. In the case of concrete, fire-induced changes arising from physio-chemical reactions often leads to degradations in physical and microstructural mechanical properties and may trigger spalling. Spalling is defined as the explosive break up of chunks of concrete from the structural member during a fire. As such, spalling not only reduces the available overall cross-sectional area in a reinforced concrete (RC) member, but also exposes steel reinforcement and internal concrete layers to fire; thus accelerating the rate of strength and modulus deterioration throughout a structural member [1]. This loss in strength and modulus properties, combined with loss of cross section, diminishes fire resistance of a RC structural members. These hostile effects can adversely impact performance of fire exposed concrete columns.

The fire resistance of a structural member is influenced by a number of factors, including fire scenarios, geometry and boundary conditions, and these factors are changing with fire exposure.

Thus, fire resistance valuation involves complex set of calculations and requires insights into varying material and structural behavior under the combined effects of high temperatures and mechanical loading. Due to the costly, lengthy and specialized nature of fire testing, researchers and engineers heavily rely on hand-based analytical or empirical methods or cumbersome finite element (FE) based numerical simulations and digital twinning, with various degrees of idealization, to design structural members to withstand fire effects [2,3]. While the aforenoted methods are regarded as exemplary procedures, it must be noted that the validity of such methods stems from prior calibration against actual fire tests. However, the serious absence on standardized validation procedure (in terms of proper selection of simulation environment, assignment of inputs etc.) adds layers of challenges that continue to hinder the use, as well as acceptance, of traditional calculation methods. In fact, a look into the open literature published over the past two decades shows that advancements in developing robust calculation methods for structural fire engineering applications continues to be minute as opposed to those observed in parallel fields (i.e. earthquake engineering [4,5]).

To accelerate research efforts and support the current inertia aimed at facilitating performance-based design, fire researchers and engineers are to fully leverage ongoing advancements in computing and data analytics. Of interest to this research area are methods that are capable of comprehending fire behavior of materials and structural systems to enable rapid and robust assessment of performance of structural members under fire conditions. Such methods are preferred to be easy to use and deploy (i.e. do not require specialized/commercial software or workstations) since structural engineers are not often familiar in fire design aspects. Further, such methods are required to be well validated, and also need to be universally accepted and unified to facilitate wide and prompt acceptance [6–9].

Given the multi-dimensionality of most fire engineering related phenomena, arriving at such methods was not truly possible until the recent advances of machine learning (ML) techniques. This technology capitalizes on the notion that machines can be trained to learn how humans (i.e. structural fire engineers) think and practice as to identify patterns and solutions to complex phenomena (e.g. spalling). In a way, these envisioned tools are "expert systems" that complement human engineers and researchers' expertise [10,11].

The use of ML has been rapidly evolving over the past few years but been slowly implemented into structural and fire engineering applications. For example, Golafshani and Behnood [12] applied a combination of ML algorithms to predict the compressive strength of various concretes with success. Similarly, Chan et al. [13] proposed and trained a basic artificial neural network (ANN) by analyzing small scale material property tests to arrive at a surrogate model to predict temperature-induced degradation in compressive strength of concrete. Erdem [14] also developed a simple ANN to estimate flexural capacity of RC slabs under fire conditions by accounting for material characteristics (e.g. compressive strength of concrete, yield strength of reinforcement) and geometric features (i.e. depth, span) etc. In a notable work, McKinney and Ali [15] developed a general purpose ANN to qualitatively classify spalling in concrete cylinders. It is worth noting that the use of other ML algorithms; such as genetic programming, to tackle fire-related problems is also reported in the literature [16–18].

A common feature of many of the above discussed works is the implied notion of *additional validation* to showcase that ML can be used to predict a phenomenon that happens to be related to fire engineering. While ML algorithms have been rigorously vetted by computer and data scientists, whether during stages of development or deployment, validating the potential of using ML in the field of fire engineering still has its merit but do not fully harness the full potential of ML. Another common notion to consider is that most of the reviewed works seem to apply ML techniques in a "black-box" approach. In this approach, a ML algorithm is but a tool that generates a surrogate model with highly complex and unclear configuration to link *inputs-to-output(s)*. Such surrogate models cannot be simplified to understand its inner workings, nor how/why such a model yields a certain prediction of response parameters [19,20].

Unlike traditional works, this paper explores the use of explainable ML (Exp-ML) to enable developing new and improved ML tools that are transparent and can truly provide valuable insights to fire researchers and engineers. This work also tackles the limited availability of fire test data by proposing the use of a combination of real, synthetic and augmented observations. For this, a ML ensemble has been calibrated and validated to predict fire resistance and spalling of RC columns subjected to either standard or design fire scenarios from one-, two-, three- and four-sides thus; covering a wide range of practical situations that may occur in real life fire incidents. The proposed approach can also be easily extended to other structural members and fire scenarios and hence paves the way to smoothly integrate Exp-ML into wide ranging structural fire engineering problems.

## 2.0 FACTORS GOVERNING FIRE RESISTANCE AND FIRE-INDUCED SPALLING

For the most part, RC columns made of traditional concrete (i.e. normal strength concrete (NSC)) exhibit excellent performance under fire conditions provided that such columns have adequate size and concrete cover [21]. Nowadays, high strength concrete (HSC) and ultra-high performance concrete (UHPC) are much prominent and attractive due to their superior mechanical and durability properties at ambient conditions. However, the same concretes are shown to exhibit poor performance under fire conditions [22,23]. Numerous studies have noted that HSC and UHPC not only undergo faster degradation in strength and modulus properties due to elevated temperatures, but are shown to be more susceptible to fire-induced spalling than NSC [24].

This relatively high susceptibility to spalling is driven by the complex microstructural changes, low permeability and extensive use of fillers/admixtures in HSC and UHPC [1,25]. These factors results in build-up of high level of vapour pressure during fire exposure – the main cause of spalling. Due to the dense microstructure of HSC and UHPC, vapour pressure build-up can reach high degree of saturation which translates into large internal pressure or tensile stresses (estimated at 5-8 MPa). Such internal pressure is often too high to be resisted by the rapidly degrading tensile strength of HSC and UHPC (about 5-7 MPa) [26]. All being equal, HSC and UHPC become brittle at elevated temperatures and the strain attained at any stress level is often much lower than that attained in NSC for any given temperature which explains the favorable performance of NSC under fire [27].

It is worth noting that observations from fire tests as well as real fire incidents have identified a series of key factors that govern fire performance and spalling susceptibility of concrete [28–31]. A brief summary of these factors is provided herein and a more comprehensive review can be found elsewhere [30,32–34].

*2.1 Concrete mixture design and properties*
The components used in concrete mixture govern the ambient and elevated temperature properties of concrete. The mechanical properties of concrete are much more affected by the rise in temperature (and temperature-induced changes) as opposed to thermal properties and hence the former are of interest to this discussion. Some of the key factors that affect fire performance of concrete include: aggregate type, silica fume, density, fiber content, compressive strength, moisture content, etc. [27,35,36].

There are three main types of aggregates commonly used in batch mixture design; carbonate, silicate and light weight aggregate. Of these types, carbonate aggregates are noted to provide better fire resistance than that of silicate and light weight aggregates. This can be attributed to the fact that carbonate aggregates undergoes an endothermic reaction at 700°C due to dissociation of the dolomite. This reaction acts as a heatsink and lowers the rate of temperature rise and slows down strength deterioration in carbonate concrete [23]. On a similar note, fire tests carried out by Bildeau et al. [37] showed that the extent of spalling was found to be greater when lightweight aggregate is used possibly due to higher moisture content contained by lightweight aggregate which creates higher vapor pressure under fire. Overall, high moisture content has been shown to increase concrete' vulnerability to spalling [38]. It should be clearly noted that special attention needs to be directed towards the initial saturation degree, as in the case of relative humidity, since dense concretes take relatively longer period to reach a moisture level (content) in equilibrium with the environment.

Another factor that falls under concrete batch mixture design is silica fume (a by-product from the production of elemental silicon or alloys containing silicon). Observations from fire tests indicate that RC columns made with silica fume are at higher risk to spalling than concrete without silica fume. Columns made with silica fume not only underwent large spalling (in excess of 15%) but also had 30% lower fire resistance than traditional columns [39]. This susceptibility to spalling can be attributed to the increased compactivity facilitated by silica fume. Densely compacted concretes have lower permeability which restricts the loss of moisture during curing, and fire testing [40].

In lieu of silica fume, synthetic or organic fibers can be added to mixture design as a means to improve concrete performance. From a fire point of view, there are two types of fibers that have been proven effective; steel and polypropylene fibers. The addition of steel fibers in the amount of about 1.75% by weight can limit spalling by enhancing tensile strength of concrete and slowing down the degradation of tensile strength with temperature rise. Polypropylene fibers can also limit spalling of concrete since these fibers melt at low temperature range (160–170°C). The melting of polypropylene fibers can create additional pores to facilitate release of vapour pressure built up under elevated temperatures exposure for overcoming spalling [41]. The addition of 0.1-0.15% by volume was recommended by various researchers [32,37].

The magnitude of concrete compressive strength can significantly influence fire-induced spalling and fire resistance of RC columns. Higher concrete strength is normally achieved through the addition of auxiliary fillers and plasticizers both of which lead to increase density and low permeability.

*2.3 Column features*
As a rule of thumb, a large sectional size positively correlates with heat and moisture transport, as well as the capacity of larger structures to store more energy, and hence large specimens are likely to spall [42]. In one study, Kanéma et al. [43] observed that spalling only occurred in large specimens as compared to smaller specimens despite all specimens being made of the same concrete mixture and were subjected to the same heating conditions. The shape of a RC column also influences its fire and spalling resistance. That being said, edged columns (i.e. of square or rectangular configuration) attract more heat due to bi-lateral transmission of heat at corners as opposed to round columns. This faster rise in sectional temperature promotes quicker degradation in mechanical properties and develops large thermal gradients; thus, promoting spalling and lower fire resistance.

The type and configuration of internal reinforcement also influence fire and spalling resistance. For example, concrete members reinforced with non-prestressed reinforcement often achieve higher fire resistance and experience less susceptibility to spalling. The poor fire resistance of prestressed reinforcement arises due to the leaner size (lower thermal mass) of prestressed members, faster degradation of strength properties in prestressing strands as compared to non-prestressed reinforcement, and denser nature of prestressed concrete. The configuration of lateral reinforcement (ties) has also been noted to affect fire resistance and fire-induced spalling of RC columns. Fire tests conducted by Kodur and McGrath [27] clearly indicate that RC columns with hooked tie configuration (bent at 135°, and with closer tie spacing) achieve improved fire performance. Using hooked ties firmly holds longitudinal rebars in place which minimizes movement and buckling of longitudinal bars and reduces the strains induced in concrete.

*2.4 Loading and heating conditions*
The magnitude and arrangement of applied loading, together with heating rate and intensity, are all factors that affect fire resistance ad spalling of RC columns. In general, fire resistance of a RC column is indirectly proportional to the magnitude and duration of applied loading since the loss of strength increases with a rise in temperature [25]. A note to remember is that eccentrically applied mechanical loadings develop a couple moment of tensile and compressive stresses on the sides of a loaded column – where tensile stresses aid in accelerating cracking of concrete and compressive forces amplifies the internal stress arising from vapour pressure. In addition to mechanical loading, heating conditions also significantly influence fire resistance and spalling of RC columns. Fires with rapid heating rates (i.e. hydrocarbon fires with 271°C/min for the first three minutes) can induce thermal shocks on a RC member by generating large thermal gradients, causing high thermal stresses as well as non-uniform expansion of exposed sides. Such gradients substantially increases the pore pressure generated in the concrete and lead to spalling [44]. While this section summaries key factors that are known to govern the spalling phenomenon, a more in depth discussion can be found in related works such as these published as part of the recent RILEM workshop [45].

**3.0 ML ENSEMBLE DEVELOPMENT**
Machine learning is a subset of artificial intelligence (AI) and aims to train machines (i.e. computing stations) to mimic the human-like reasoning process. The goal of ML is to leverage the large computing capacities of workstations to solve complex problems that may not be properly tackled via traditional methods (i.e. testing, analytical or FE simulation) or those which would require specialized software or extended simulation processing. ML differs from traditional analysis methods by its capability to identify hidden patterns in existing observations (which in the context of this work refer to outcomes from fire tests etc.) by leveraging free-form and nonparametric evolutionary algorithms that do not require predefined set of assumptions or idealization to search for solutions (as opposed to mathematical or statistical approaches) [46].

### *3.1 Rationale*

The rationale behind utilizing ML in this study stems from the understanding that there exists at least one governing relation between fire-related phenomena (e.g. fire resistance and/or fire-induced spalling) of RC columns with the key factors identified in Sec. 2. This relation can be arrived at through a rigorous investigation of observations from a large number of tests carried out carried out to evaluate fire resistance and fire-induced spalling in RC columns. Given that both phenomena entitle multi-dimensional factors implies that arriving at the so-called relation(s) is effectively complex and will require thorough analysis to realize. However, this analysis can somewhat be easily achieved if one to use modern methods of higher order than that associated with traditional approaches. Through logical understanding of a fire resistance and spalling phenomena, and not just through satisfying numerical objectives often achieved through traditional analysis, ML rises as a potential candidate.

Oftentimes, the use of one ML algorithm to understand a phenomenon can be sufficient. However, recent experience have shown that this practice might lead to develop biased ML-based solutions in some situations or in some instances may not yield to achieving an optimal solution in a timely manner [47,48]. As such, the current work explores the notion of ensemble learning. This type of ML harnesses the advantages of multi-algorithm search; thus, enabling rapid analysis that highly correlates with attaining the most optimal solution [49]. In an ensemble, a number of ML algorithms can search in harmony or in competitive arrangement to look for optimum solutions. Once a solution is identified by each algorithm, a series of fitness metrics are applied to identify the fittest solution for a problem (say expected fire resistance duration of a RC column) [17]. Following this procedure, the identified solution is not only vetted across different search mechanisms but is also vetted through different analysis stages (see Fig. 1).

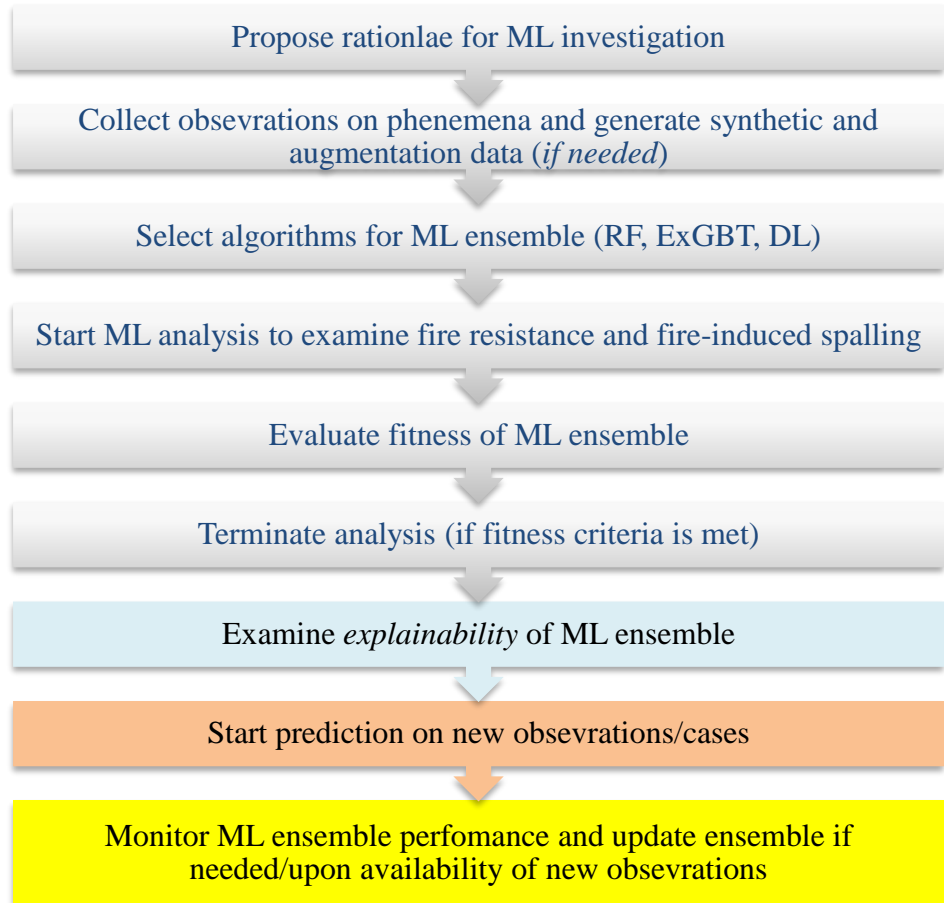

Fig. 1 Flowchart of developed ML procedure for evaluating fire resistance of and spalling of RC columns

In some scenarios, an ensemble may look for a qualitative solution to a problem. For example, a fire engineer may seek an answer to the following question; *will a RC column with particular feature spalls if exposed to fire or not?* In this scenario, each algorithm in the developed ensemble will generate a "yes/no" answer (i.e. algorithm A: the column is expected to spall, algorithm B: the column is not expected to spall etc.). Hence, the ensemble reviews all answers to arrive at a decision in a similar manner to "majority voting" where the final answer of the ensemble is given as that observed by the majority of the participating algorithms.

In all cases, whether a ML ensemble arrives at a numeric prediction (e.g. fire resistance in minutes) or categorial prediction (e.g. spalls/does not spall), the fire engineer needs to understand how such a decision is made. This is where explainable ML (Ex-ML) shines as such a technique is primed to provide complete insights into ensemble working process as opposed to the more traditional "black-box" ML approach. This explainability notion can also be used to understand how key factors (such as those listed in Sec. 2) interact with each other as well as with the outcome of a given phenomenon.

Both of the above discussed points are not currently attainable using traditional fire resistance analysis approaches. For example, a fire engineer can trace stress development within a FE model throughout a FE simulation. However, any insights from such analysis will always be constrained by the input variables used in the FE analysis which may not be in close proximity to those used in the fire tests or in those used constructions (i.e. for post-fire investigation). This practice becomes problematic when FE models are extended beyond their original cases. Given that we still lack standardized simulation procedure that is free of limiting assumptions and account for complex phenomenon associated with modern construction materials (such as spalling in HSC) casts grey shadows over the validity of extending FE models. Given that ML utilizes a vast amount of data implies that an ML analysis is capable of better representing the actual behavior of materials and structural members as opposed to FE models.

Once the ML ensemble is trained and validated, then this ensemble can be applied to predict fire resistance and possibility of fire-induced spalling of RC columns. All that is needed is to input the selected features into the developed ensemble to realize an outcome. Over time, it is a good practice to monitor the prediction capability of the ensemble as the predictivity of some ML ensembles might decay/worsen. Once additional observations are collected, the ML ensemble can be re-trained to improve its prediction capability and ensure that it is updated with advancements in concrete material technology (i.e. development of new concrete mixtures etc.) [50].

### 3.2 Database development and proposed techniques for synthetic and augmentation data generation

As the case of most ML-based analysis, a comprehensive review of literature is to be carried out to compile a database of real observations taken in fire resistance tests. In this study, two fire-related phenomena are to be examined via ML including fire resistance and spalling of RC columns. The conducted literature review has identified 167 fire tests taken on various RC columns spanning over three decades of research [31,32,51–62]. Common features in all columns were then collected. The identified features were selected to align with recommendations of earlier studies (discussed in Sec. 2) and to enable developing an Exp-ML ensemble that is useful for both, fire designers and researchers [63–65].

The identified critical parameters include: 1) column width, $W$, 2) steel reinforcement ratio, $r$, 3) length, $L$, 4) concrete compressive strength, $f_c$, 5) steel yield strength, $f_y$, 6) restraint conditions, $K$ (fixed-fixed, fixed-pinned, and pinned-pinned), 7) concrete cover to reinforcement, $C$, 8) eccentricity in applied loading in two axes ($e_x$ and $e_y$), 9) magnitude of applied loading, $P$, 10) fire exposure scenario, $E$ (ranging from ASTM E119, hydrocarbon, design fires etc.), 11) number of exposed faces, $S$ (1-, 2-, 3- and 4- faces), 12) fire resistance (of failure time), $FR$, and 13) Spalling, $SP$ (Yes, or No). One must note that results on spalling were available on 167 RC columns, while only results from 144 test were available for fire resistance of RC columns.

Noting how observations from fire tests are scarce and limited, this work proposes the use of synthetic and augmented observations. Synthetic observations are those defined to be generated by manipulation of real observations. Despite such manipulations, synthetic observations maintain the same schema and statistical properties as its real counterparts. Synthetic observations can be obtained via a number of procedures such as Synthetic Minority Over-sampling Technique (SMOTE) and Modified-SMOTE [66]. The SMOTE procedure creates new observations which lie between any two nearest real observations joined by a straight line and then calculates the distance between two real observations in the feature space, multiplies the distance by a random number between 0 and 1 and places the newly generated synthetic observation at this new distance from one of the real observations used for distance calculation. One should note that this technique has been proven effective in a variety of machine learning problems and hence is explored herein [67].

Thus, a similar procedure to SMOTE was employed herein. In this procedure real RC columns with similar features were clustered in pairs. Then, observations from fire tests for each RC column in each pair were compared. If both columns have spalled (or not spalled), then a third synthetic column is generated with features similar to those in the two parent real columns. The newly generated column is said to have averaged features to the real ones and a similar observation as well (i.e. if the real columns have spalled, then the synthetic column is also expected to spall since it has averaged features to the real ones). In case that two columns in a pair have different spalling observation (e.g. one column spalled, while the other did not), then the synthetic column is assumed to have the worst case scenario (for conservativeness). Based on this procedure, 166 synthetic columns and observations were added to the 167 real observations (fire tests on columns).

The second technique that can be used to enlarge the size of a database is to use augmented observations. Augmented observations are those obtained by means of advanced analysis methods such as validated FE models. It is worth mentioning that the use of augmented observations from various studies/investigations, as opposed to one source, overcomes many of the aforenoted limitations of FE simulations. For example, using augmented results from European researchers who are likely to apply Eurocode material models to their FE models, in addition to North American researchers who are likely to used ASCE material models in their FE models, provides a variety of observations that can come in handy to expand a fire database. In this work, 160 augmented observations were collected from the open literature [68–72]. After all, the compiled database contained 494 RC columns.

Figure 2 and Table 1 lists additional details into the range of each of the selected features in the collected RC columns. This table also provides insights into the range of applicability of the developed ensemble and shows the close resemblance between real, synthetic and augmented observations; thus, providing a layer of confidence into the proposed techniques. One should note that the above features were available for all collected RC columns. Still, other features can also be included as all that is needed is to collect information on a new feature. For example, humidity of concrete was not selected herein since it was only reported in a few studies (and not all studies) to maintain homogeneity of the compiled database. Further, this factor is not readily available to designers most of the time and hence may not be a viable input to the fire analysis via ML. Previous works provided solutions on tackling similar challenges [73,74].

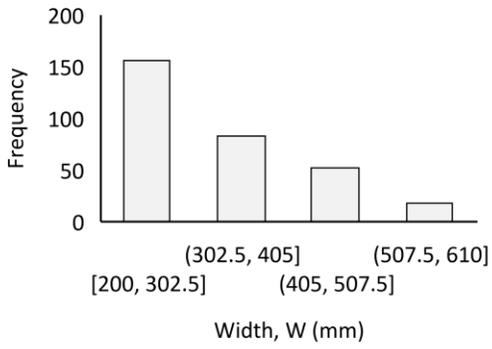
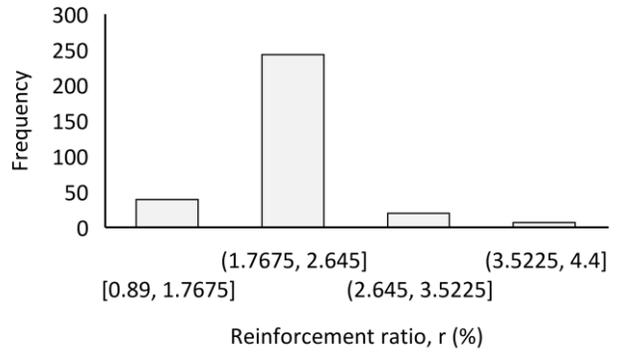
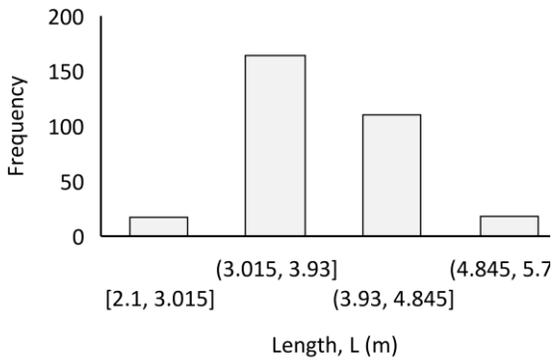
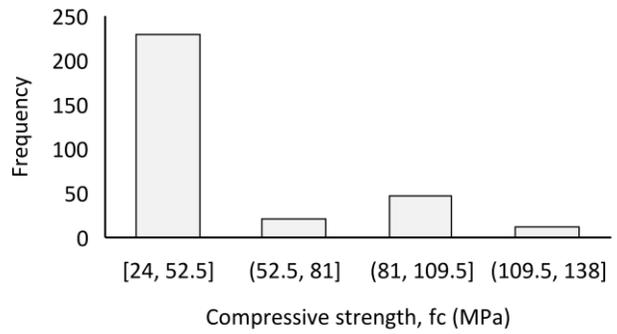
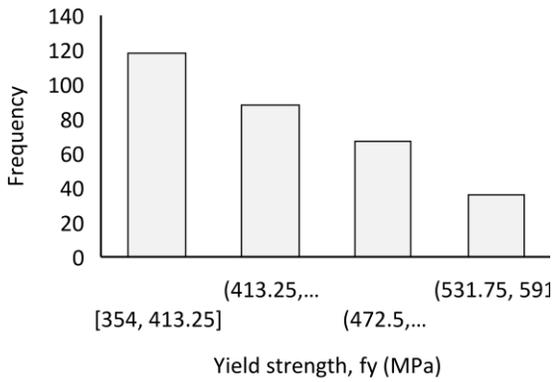
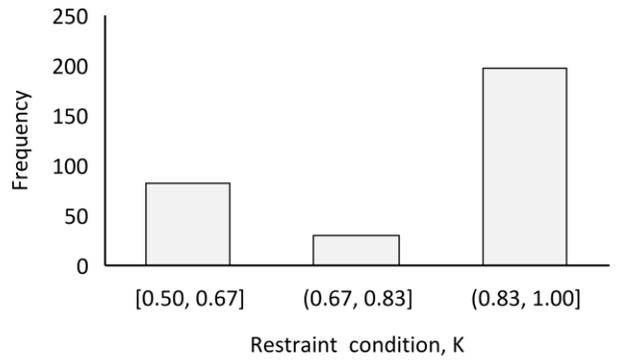
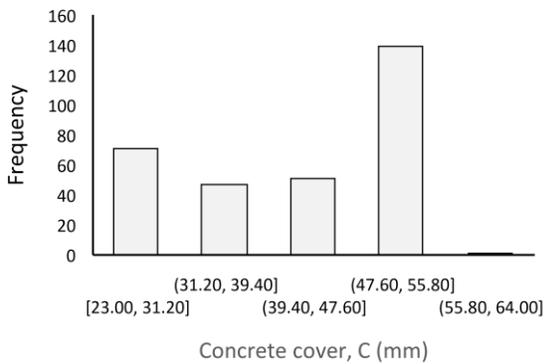
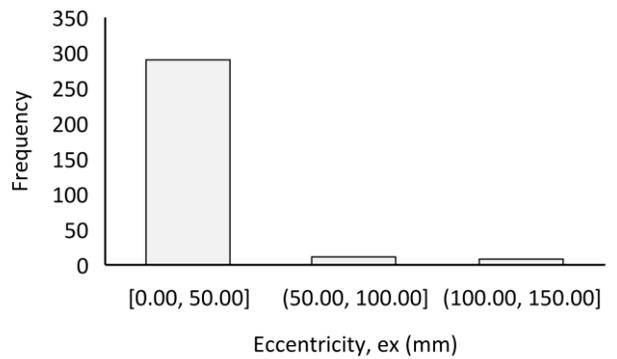

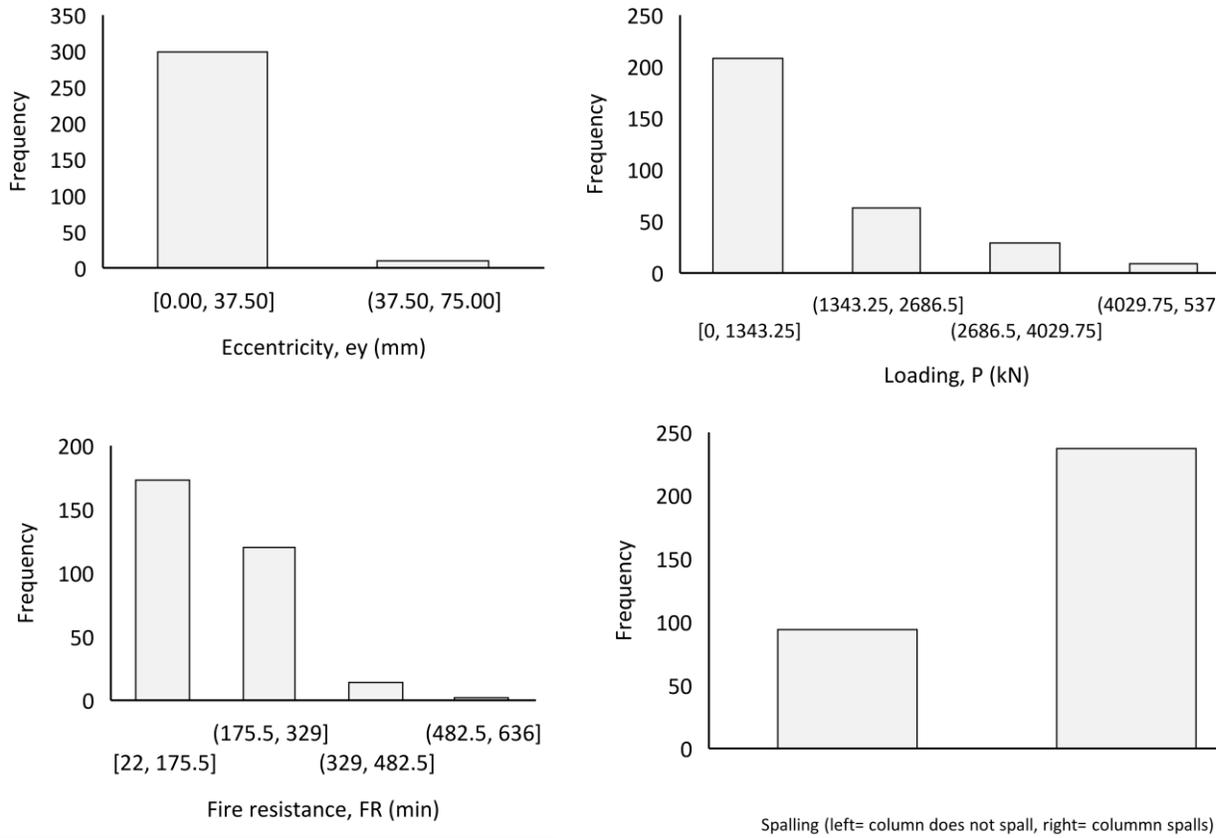

Fig. 2 Frequency of identified features of selected RC columns in the compiled database

Table 1 Statistics on collected database

| | | | W (mm) | r (%) | L (m) | $f_c$ (MPa) | $f_y$ (MPa) | C (mm) | $e_x$ (mm) | $e_y$ (mm) | P (kN) | FR (min) |
|---|---|---|---|---|---|---|---|---|---|---|---|---|
| Fire resistance analysis | Real observations | Minimum | 203.0 | 0.9 | 2.1 | 24.0 | 354.0 | 25.0 | 0.0 | 0.0 | 0.0 | 55.0 |
| | | Maximum | 610.0 | 4.4 | 5.7 | 138.0 | 591.0 | 64.0 | 150.0 | 75.0 | 5373.0 | 389.0 |
| | | Average | 350.4 | 2.1 | 3.9 | 55.7 | 439.3 | 42.4 | 12.8 | 3.2 | 1501.8 | 176.6 |
| | | Standard deviation | 105.3 | 0.5 | 0.5 | 33.0 | 61.8 | 7.1 | 23.3 | 12.9 | 1168.6 | 82.0 |
| | | Skewness | 1.1 | 1.0 | -0.5 | 0.9 | 1.0 | -1.0 | 3.0 | 4.0 | 1.3 | 0.4 |
| | Augmented observations | Minimum | 200.0 | 0.9 | 2.1 | 24.4 | 354.0 | 23.0 | 0.0 | 0.0 | 122.0 | 22.0 |
| | | Maximum | 610.0 | 4.4 | 5.8 | 100.0 | 576.0 | 50.0 | 150.0 | 75.0 | 3190.0 | 391.0 |
| | | Average | 333.9 | 2.1 | 4.1 | 48.1 | 436.3 | 41.2 | 16.1 | 3.7 | 1108.1 | 155.9 |
| | | Standard deviation | 117.2 | 0.5 | 0.6 | 26.4 | 52.2 | 9.9 | 30.2 | 13.6 | 874.5 | 81.8 |
| | | Skewness | 1.2 | 0.9 | 0.5 | 1.3 | 1.2 | -0.8 | 2.7 | 3.7 | 1.3 | 0.2 |
| | All observations | Minimum | 200.0 | 0.9 | 2.1 | 24.0 | 354.0 | 23.0 | 0.0 | 0.0 | 0.0 | 22.0 |
| | | Maximum | 914.0 | 4.4 | 5.8 | 138.0 | 591.0 | 64.0 | 150.0 | 75.0 | 5373.0 | 636.0 |
| | | Average | 324.3 | 2.1 | 4.0 | 49.3 | 449.4 | 40.2 | 15.8 | 2.0 | 1204.8 | 161.0 |
| | | Standard deviation | 99.2 | 0.6 | 0.7 | 28.1 | 60.1 | 8.7 | 29.7 | 10.1 | 1031.6 | 97.6 |
| | | Skewness | 1.9 | 0.6 | 0.3 | 1.4 | 0.7 | -0.6 | 2.9 | 5.3 | 1.7 | 0.9 |
| Spalling Analysis | Real observations | Minimum | 152.0 | 0.3 | - | 15.0 | - | 13.0 | - | - | 0.0 | - |
| | | Maximum | 514.0 | 11.7 | - | 126.5 | - | 64.0 | - | - | 5373.0 | - |
| | | Average | 326.1 | 2.3 | - | 42.1 | - | 33.5 | - | - | 1342.0 | - |
| | | Standard deviation | 71.8 | 1.6 | - | 24.3 | - | 7.7 | - | - | 1001.3 | - |
| | | Skewness | 0.8 | 2.5 | - | 1.9 | - | -0.4 | - | - | 1.6 | - |
| | Synthetic observations | Minimum | 152.0 | 0.5 | - | 16.1 | - | 13.0 | - | - | 84.5 | - |
| | | Maximum | 511.0 | 10.8 | - | 119.7 | - | 51.0 | - | - | 4970.0 | - |
| | | Average | 325.6 | 2.3 | - | 42.0 | - | 33.5 | - | - | 1338.4 | - |
| | | Standard deviation | 66.0 | 1.3 | - | 23.1 | - | 7.3 | - | - | 911.8 | - |
| | | Skewness | 0.8 | 2.5 | - | 1.9 | - | -0.7 | - | - | 1.6 | - |
| | All observations | Minimum | 152.0 | 0.7 | - | 16.0 | - | 25.0 | - | - | 0.0 | - |
| | | Maximum | 514.0 | 4.9 | - | 126.5 | - | 64.0 | - | - | 5373.0 | - |
| | | Average | 325.3 | 2.5 | - | 54.3 | - | 37.6 | - | - | 1556.9 | - |
| | | Standard deviation | 69.4 | 0.8 | - | 27.9 | - | 4.4 | - | - | 1109.1 | - |
| | | Skewness | 0.7 | 1.0 | - | 1.1 | - | 0.6 | - | - | 1.4 | - |

### *3.4 Ensemble development*

Fire resistance analysis through ML is quite different than that using traditional calculation approaches. Traditional approaches necessitate the development of two sets of models. In the first set, the thermal response of a RC column is arrived at. Temperature rise and propagation within the column are obtained and then are inputted into the second model set, in addition to mechanical loading, to evaluate structural response under fire conditions. Both steps of analysis can be performed at sectional or member level via a generic or a special finite element package. To enable such analysis, proper inputs such as temperature-dependent material properties, together with meticulous consideration to meshing, convergence criteria etc. are required [75,76].

Unlike traditional analysis, a ML analysis can evaluate fire resistance and/or spalling tendency instantaneously and without the need to compile idealized input parameters. This is true knowing that since the outcome of fire tests (i.e. spalling or fire resistance) is known, then a properly developed ML model can relate the aforenoted and identified key features to the outcome of fire tests while implicitly accounting for temperature-dependent properties of concrete and steel and negating any need for idealization or complex model building/preparation as well as overcome convergence issues to reach a solution. For a ML ensemble analysis to start, a user must select a series of ML algorithms. The selection process can be purely arbitrary or can be taken as a result of a sensitivity analysis [77].

In this work, a combination of ML algorithms was examined first. This examination process aimed to identify a suitable ensemble with balanced accuracy and search/deployment speed. The selected ensemble is identified as one that contains the following algorithms: random forest (RF), extreme gradient boosted trees (ExGBT), and deep learning (DL). A brief description of these algorithms is provided herein and additional details can be found elsewhere [78–81].

The RF algorithm randomly generates multiple decision trees (and hence the term "random forest") to analyze a phenomenon [78]. This algorithm is a nonparametric classifier that analyzes the outcome of each individual decision tree to reach a predictive outcome. In a classification problem such as that tackling the tendency of a RC column to spall or to not spall, the *majority voting* method is used to arrive at the final output; such that:

$$Y = \frac{1}{J}\sum_{j=1}^{J} C_{j,full} + \sum_{k=1}^{K}\left(\frac{1}{J}\sum_{j=1}^{J} contribution_j(x,k)\right) \tag{1}$$

where, $J$ is the number of trees in the forest, $k$ represents a feature in the observation, $K$ is the total number of features, $c_{full}$ is the average of the entire dataset (initial node). The developed ExGBT also employed a maximum tree depth of 10, learning rate of 10% and 500 stages of boosting.

The second algorithm is extreme gradient boosted trees (ExGBT) which is a special form of the Adaboost algorithm developed by Freund and Schapire [79]. ExGBT re-samples the collected observations into decision trees, where each tree sees a boostrap sample of the database in each iteration. ExGBT shares some similarity with RF except that ExGBT does not fits decision trees in parallel, but rather fits each successive tree to the residual errors from all the previous trees combined. As a result, ExGBT focuses each iteration on the observations that are most difficult to predict; which becomes a good practice for the algorithm to yield high prediction accuracy (see Eq. 2).

$$Y = \sum_{k=1}^{M} f_k(x_i), f_k \in F = \{f_x = w_{q(x)}, q: R^p \to T, w \in R^T\} \tag{2}$$

where, $M$ is additive functions, $T$ is the number of leaves in the tree, $w$ is a leaf weights vector, $w_i$ is a score on $i$-th leaf, and $q(x)$ represents the structure of each tree that maps an observation to the corresponding leaf index [80]. The RF algorithm incorporated 50 leaf nodes, with a minimum of 5 samples to split an internal node.

Deep learning algorithm refers to a multi-layered artificial neural network (ANN). This algorithm mimics the neural topology of the brain. DL starts with the input layer which receives input variables (i.e. features of observations). This layer is connected to a series of hidden layers. These layers are often connected via nonlinear activation functions e.g. *Logistic*, *PReLu*, etc. that can generate an approximation form that permits gradient-based optimization etc. The outcome of this optimization process is then displayed in the last (output) layer [81]. In a way, DL aims to achieve a general, and primarily implicit, representation that best exemplifies a phenomenon (i.e. spalling); such that:

$$net_j = \sum_{i=1}^{n} In_i w_{ij} + b_j \tag{2}$$

$$Y = f(net_j) \tag{3}$$

where, $In_i$ and $b_j$ are the $i$th input signal and the bias value of $j$th neuron, respectively, $w_{ij}$ is the connecting weight between $i$th input signal and $j$th neuron and $f$ is a *PReLu* activation function. The number of used layers are 64, with 3% learning rate, and *Adam* optimizer to enhance processing of observations.

## 4.0 PERFORMANCE AND INSIGHTS INTO EXP-ML ENSEMBLE

The developed ML ensemble was trained on 70% of the complied database and tested against 30% of the remaining RC columns. This 70/30 ratio split was arrived as part of a preliminary analysis to identify the most accurate split after a series of combinations of recommended splits were tried [62]. This section provides details on the performance of the developed ensemble, together with insights on both fire phenomena examined herein.

### *4.1 Fire-induced spalling analysis*

The performance of the ensemble in correctly predicting spalling of RC columns can be evaluated through a look at its performance metrics. Primarily, three performance metrics were used for the spalling phenomenon, Logarithmic Loss (i.e. log Loss), area under the receiver operating characteristic (ROC) curve and confusion matrix. The log loss metric measures the performance of a classification ML ensemble whose output is a probability value between 0 and 1. A perfect ML ensemble would have a log loss of zero. In a binary classification such as that tackled here, the log loss can be estimated as:

$$-(y \log(p) + (1 - y) \log(1 - p)) \tag{4}$$

where, $y$ and $p$ are actual and predicted observations.

The ROC curve on the other hand is a graphical plot that illustrates the predictive capability of a binary ensemble as its discrimination threshold is varied receiver operating characteristic [82]. Since ROC is a probability curve, then the area under this curve measures the degree of separability. For a perfect ensemble, the area under this curve should be equal to unity. The performance of the developed ensemble is 0.35(0.36) and 0.91(0.90) for logloss error and area under the curve, respectively for training and (validation) stages. Other metrics from confusion matrix were also investigated. For example, the ensemble reached: sensitivity (95%), fallout (36%), specificity (63%), precision (86%), and accuracy (86%). The above clearly implies that the developed model is quite accurate given the complexity and randomness of spalling occurrence as a phenomenon. It is worth noting that the developed ensemble is capable of accurately analyzing 350 RC columns for spalling per second; thus, significantly accelerate fire design in our practice.

Table 2 shows the results of feature association (i.e. mutual information) and feature correlation between the selected features and fire-induced spalling. This matrix provides insights into association strength between pairs of features which shows the extent to which features depend on each other where large values indicate serious collinearity between the features involved. This matrix shows that all selected features have positive dependence with fire-induced spalling (with values close to 1.0 imply higher association). In more details, $r$, $f$, $b$, $P$, and $C$ has the strongest *individual* dependence with fire-induced spalling (in this order). It is worth noting that $b$ and $r$ and $C$ and $r$ also have strongly association with other. The same table also shows the selected features with positive or negative correlation with fire-induced spalling. One should note that a correlation of ±1.0 implies perfect correlation, 0.5-1.0 (strong), 0.3-0.5 (moderate), 0.3-0.1 (weak), 0-0.1 (none). In more details, $f$, $b$, and $P$ has the highest positive *individual* correlation with fire-induced spalling. It is worth noting that $P$ and $f$ and $b$ and $P$ also have strongly correlated with other.

Table 2 Feature analysis between selected features and spalling

| | Association matrix | | | | | |
|---|---|---|---|---|---|---|
| | SP | $r$ | $f$ | $b$ | $P$ | $C$ |
| SP | 1.00 | 0.52 | 0.45 | 0.41 | 0.33 | 0.34 |
| $r$ | | 1.00 | 0.31 | **0.60** | 0.34 | **0.57** |
| $f$ | | | 1.00 | 0.340 | 0.25 | 0.34 |
| $b$ | | | | 1.00 | 0.41 | 0.45 |
| $P$ | | | | | 1.00 | 0.39 |
| $C$ | | | | | | 1.00 |
| | Correlation matrix | | | | | |
| | $b$ | $r$ | $f$ | $C$ | $P$ | SP |
| $b$ | 1.00 | | | | | |
| $r$ | -0.19 | 1.00 | | | | |
| $f$ | 0.15 | -0.01 | 1.00 | | | |
| $C$ | -0.02 | 0.21 | 0.40 | 1.00 | | |
| $P$ | **0.60** | 0.07 | **0.65** | 0.34 | 1.00 | |
| SP | 0.28 | -0.06 | 0.14 | -0.19 | 0.20 | 1.00 |

On a similar note, the ML ensemble shows the different impact of the selected features by taking their interaction into account as opposed to their individual dependence to the spalling phenomenon. This analysis identifies which features are driving model predictions the most using the SHAP (SHapley Additive exPlanations) technique [83] to explain individual predictions. This analysis infers that $C$ (100%), $f$ (95%), $b$ (88%), $r$ (66%) and $P$ (44%) are the most impactful features when it comes to spalling. From this, one can see the main three features that can significantly affect spalling of RC columns are $C$, $f$ and $b$. Figure 3 shares additional insights into the impact of each of these features on the increased possibility of spalling (when all other features remain constant). It is quite clear that increasing cross sectional size and compressive strength of concrete is associated with higher susceptibility to spalling while increasing concrete cover seems to limit spalling.

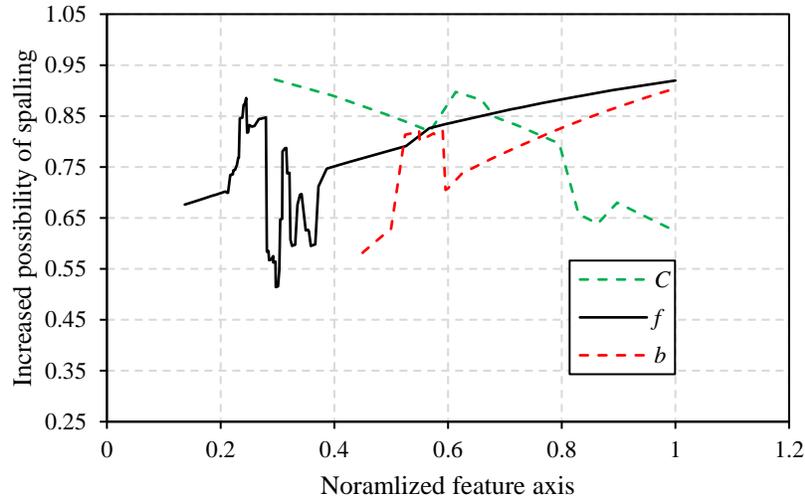

Fig. 3 Insights into key factors influencing spalling in RC columns

*4.2 Fire resistance analysis*

The fire resistance analysis carried out herein comprised of two approaches. In the first approach, a RC column can be classified to belong to a class that falls under one, two, three, or four hour fire rating. A more refined approach was also developed in which a RC column is analyzed to evaluate its fire resistance in minutes. Both approaches are un-coupled and can be run in parallel to arrive at fire resistance. The reasoning behind developing two approaches is to provide fire researchers and designers with convenient tool that may fit in different applications.

The prediction accuracy of the developed ensemble was then evaluated twice (once for each approach). In the first approach, the ensemble was not able to achieve as high of metrics as that in the spalling case. The ensemble achieved a relatively high log loss error of 0.87 and 0.98 for validation and testing, respectively, but managed to realize comparatively good area under the ROC curve of 0.88 and 0.84 for the aforenoted stages. A key point to remember is that some of the used RC columns failed abruptly in under 60 min (especially those collected from Hass tests [84] and seemed to adversely affect the ML analysis. Further investigation is currently underway to identify possible solutions to overcome this issue and salvage these columns, as opposed to simply neglecting such columns from the ML analysis. As such, the discussion on this approach will be visited in a future work.

On a more positive note, the performance of the ensemble was much improved when it came to the second approach for fire resistance analysis. For this approach, more appropriate metrics are to be applied as the analysis shifts from classification into regression. These metrics were, correlation coefficient ($R$), coefficient of Determination ($R^2$), and Root Mean Squared Error ($RMSE$) – See Eq. 5-7. These metrics comes to 0.89(0.86), 0.80(0.74), and 45.59(41.49) for training and testing, respectively.

$$R = \frac{\sum_{i=1}^{n}(A_i-\bar{A}_i)(P_i-\bar{P}_i)}{\sqrt{\sum_{i=1}^{n}(A_i-\bar{A}_i)^2 \sum_{i=1}^{n}(P_i-\bar{P}_i)^2}} \quad (5)$$

$$R^2 = 1 - \sum_{i=1}^{n}(P_i - A_i)^2 / \sum_{i=1}^{n}(A_i - \bar{A}_i)^2 \quad (6)$$

$$RMSE = \sqrt{\frac{\sum_{i=1}^{n} E_i^2}{n}} \quad (7)$$

where, $E$ is the error between actual observation ($A$) and predicted observations ($P$). The bar sign refers to the average of observations.

It can be seen from the above validation that the above ensemble can be used with confidence to further examine the phenomenon of fire resistance of RC columns. Table 3 shows the outcome of association and correlation analysis between the identified features and fire resistance (FR) of RC columns. The first row of this matrix signifies the individual dependence between all features and FR. As can be seen in the table, all features seem to have a relatively comparable dependence except for $S$, $e_y$ and $E$. The same matrix also shows that there is strong association between the majority of features and $L$, $f_y$, and $C$. Table 3 also shows the correlation matrix of identified features and fire resistance of RC columns. The bottom row of this matrix signifies the individual dependence between all features and FR. As one can see, all features seem to have a relatively comparable correlation except for $r$, $e_y$, $S$ and $E$. The same matrix also shows that there is varying degree of correlation between the majority of features $b$, $K$, $C$, $L$, $f_y$, and $C$.

Table 3 Feature analysis between selected features and fire resistance

| | | | | | | Association matrix | | | | | | | |
|---|---|---|---|---|---|---|---|---|---|---|---|---|---|
| | FR | L | C | b | $f_y$ | r | $e_x$ | P | K | $f_c$ | S | $e_y$ | E |
| FR | 1.00 | 0.29 | 0.27 | 0.26 | 0.29 | 0.31 | 0.23 | 0.27 | 0.17 | 0.25 | 0.03 | 0.06 | 0.10 |
| L | | 1.00 | 0.55 | 0.51 | 0.63 | 0.62 | 0.37 | 0.40 | 0.36 | 0.48 | 0.02 | 0.09 | 0.20 |
| C | | | 1.00 | 0.39 | 0.49 | 0.60 | 0.30 | 0.20 | 0.15 | 0.36 | 0.01 | 0.05 | 0.13 |
| b | | | | 1.00 | 0.46 | 0.48 | 0.25 | 0.42 | 0.18 | 0.36 | 0.03 | 0.12 | 0.18 |
| $f_y$ | | | | | 1.00 | 0.59 | 0.35 | 0.38 | 0.33 | 0.53 | 0.01 | 0.06 | 0.13 |
| r | | | | | | 1.00 | 0.26 | 0.39 | 0.31 | 0.46 | 0.02 | 0.07 | 0.17 |
| $e_x$ | | | | | | | 1.00 | 0.28 | 0.17 | 0.25 | 0.05 | 0.17 | 0.09 |
| P | | | | | | | | 1.00 | 0.17 | 0.38 | 0.02 | 0.07 | 0.13 |
| K | | | | | | | | | 1.00 | 0.21 | 0.01 | 0.04 | 0.04 |
| $f_c$ | | | | | | | | | | 1.00 | 0.02 | 0.11 | 0.13 |
| S | | | | | | | | | | | 1.00 | 0.20 | 0.00 |
| $e_y$ | | | | | | | | | | | | 1.00 | 0.01 |
| E | | | | | | | | | | | | | 1.00 |
| | | | | | | Correlation matrix | | | | | | | |
| | b | r | L | $f_c$ | $f_y$ | K | C | $e_x$ | $e_y$ | P | E | S | FR |
| b | 1.00 | | | | | | | | | | | | |
| r | -0.12 | 1.00 | | | | | | | | | | | |
| L | -0.17 | 0.26 | 1.00 | | | | | | | | | | |
| $f_c$ | 0.24 | 0.06 | -0.11 | 1.00 | | | | | | | | | |
| $f_y$ | -0.25 | -0.35 | -0.08 | -0.48 | 1.00 | | | | | | | | |
| K | 0.02 | -0.28 | 0.33 | -0.08 | 0.17 | 1.00 | | | | | | | |
| C | 0.32 | 0.31 | -0.22 | 0.28 | -0.64 | -0.36 | 1.00 | | | | | | |
| $e_x$ | -0.09 | 0.05 | 0.36 | -0.23 | 0.15 | 0.28 | -0.26 | 1.00 | | | | | |
| $e_y$ | 0.16 | -0.05 | 0.00 | -0.14 | -0.14 | 0.15 | 0.16 | 0.18 | 1.00 | | | | |
| P | 0.67 | 0.12 | -0.21 | 0.56 | -0.38 | -0.21 | 0.28 | -0.21 | 0.03 | 1.00 | | | |
| E | -0.01 | 0.00 | 0.01 | -0.03 | 0.07 | 0.02 | -0.08 | 0.04 | 0.02 | -0.01 | 1.00 | | |
| S | -0.05 | 0.02 | -0.05 | -0.07 | -0.17 | -0.17 | 0.19 | -0.04 | 0.17 | -0.03 | -0.17 | 1.00 | |
| FR | 0.38 | 0.08 | -0.44 | 0.22 | -0.28 | -0.60 | 0.56 | -0.37 | -0.04 | 0.36 | -0.05 | 0.04 | 1.00 |

The ML ensemble also provides insights into the different impact of the selected parameters by taking their interaction into account as opposed to their individual dependence to the fire resistance phenomenon using SHAP. This analysis infers that $C$ (100%), $P$ (63%), $K$ (54%), $e_x$ (52%) and $b$ (39%) are the most impactful features when it comes to fire resistance. Surprisingly, $f_c$, $r$, $L$ and $f_y$ did not rank among the top five features possibly due to the fact that such features had imbalanced frequency as shown in Fig. 2. A closer look into the highly impactful features shows that they belong two main factor groups, loading and geometric features. Figure 4 shares additional insights into the impact of each of these features on the increased possibility of improved fire resistance (when all other features remain constant). For example, eccentrically loaded columns are expected to fail much earlier than those under concentric loading resulting from high stresses due to uni- or bi-directional loading. Similarly, bigger columns are also expected to have larger fire resistance resulting from higher thermal inertia.

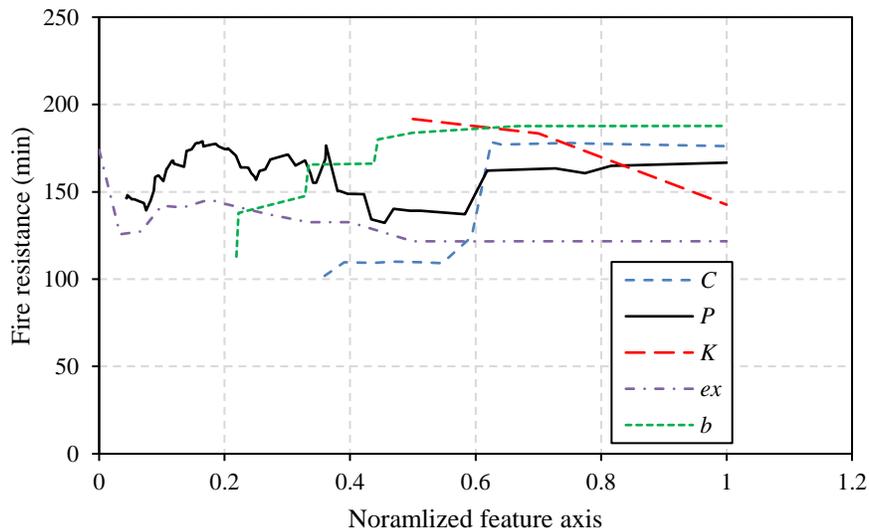

Fig. 4 Insights into key factors influencing fire resistance of RC columns

In order to further highlight the accuracy of the developed ensemble, fire resistance predictions obtained herein are also compared against codal provisions adopted by Eurocode 2 [85], and AS3600 [86] and this comparison is plotted in Fig. 5. This figures infers that predictions from the ensemble agrees well with measured fire resistance observed in fire tests, especially for those exceeding 4 hours of fire exposure. On the other hand, predictions obtained from codal methods seem to struggle to accurately predicting fire resistance in RC columns. One should still note the adequacy of Eurocode 2 predictions for columns within the 60-240 minute range beyond which these predictions seem to be underestimated. It is worth noting that both codal methods primarily account for standard fire exposure and may entail lengthy procedure. On the other hand, the developed ensemble is capable of analyzing fire resistance under various conditions many of which are not covered by codal procedure for 85 RC columns per second.

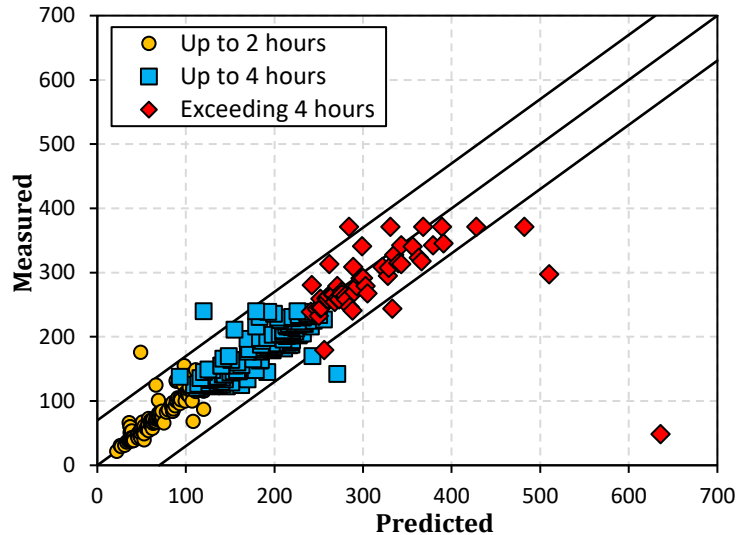

(a) Predictions from ensemble

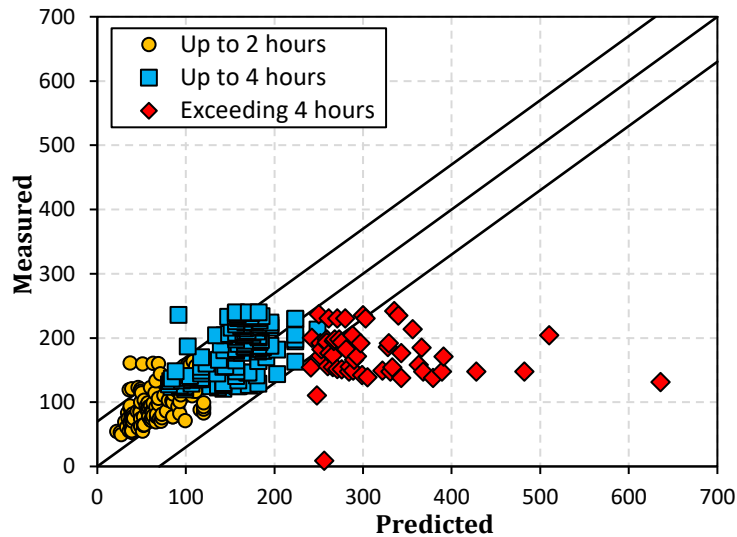

(b) Eurocode 2 method

$$R = 120\left(\frac{R_{fi}+R_a+R_l+R_b+R_n}{120}\right)^{1.8}, \text{ and } R_{fi} = 83\left(1 - \mu_{fi}\frac{1+\omega}{\frac{0.85}{\alpha_{cc}}+\omega}\right), \omega = \frac{A_s f_{yd}}{A_c f_{cd}}$$

where,
$R$ = fire resistance of column (min),
$\alpha_{cc}$ = coefficient for compressive strength,
$R_a = 1.6(a-30)$; $a$ is the axis distance to the longitudinal steel bars (mm); 25 mm $\leq a \leq$ 80 mm,
$R_l = 9.6(5-l_{o,fi})$; $l_{o,fi}$ is the effective length of the column under fire conditions; 2 m $\leq l_{o,fi} \leq$ 6 m; values corresponding to $l_{o,fi} = 2$ m give safe results for columns with $l_{o,fi} < 2$ m,
$R_b = 0.09b'$; $b' = A_c/(b+h)$ for rectangular cross-sections or the diameter of circular cross sections,
$R_n = 0$, if 4 rebars are used, and *12* for more than 4 rebars.
Performance metrics: $R = 0.7$, $R^2 = 0.49$

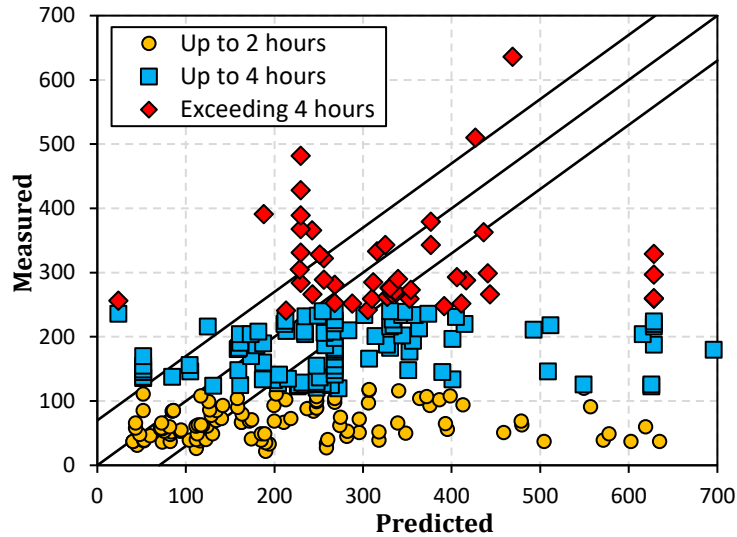

(c) AS3600 method

$$R = \frac{k \times f_c^{1.3} \times B^{3.3} D^{1.8}}{N^5 \times N^{1.5} \times L_e^{0.9}}$$

where,
$R$ = fire resistance of column (min),
$k$ = a constant dependent on cover and steel reinforcement ratio (equals to 1.47 and 1.48 for a cover less than 35 mm and greater than or equal to 35 mm, respectively),
$f_c$ = 28-day compressive strength of concrete (MPa),
$B$ = least dimension of column (mm),
$D$ = greatest dimension of column (mm),
$N$ = axial load during fire (kN),
$L_e$ = effective length (mm)
Performance metrics: $R = 0.22$, $R^2 = 0.05$

Fig. 5 Comparison of fire resistance prediction in RC columns using different approaches

## 5.0 LIMITATIONS AND FUTURE IMPROVEMENTS

One should keep in mind that an understanding of ML shows that ML algorithms analyze observations as pure data points with limited regard to their physical meaning. As such, a designer should always strive at developing explainable ML algorithms. To overcome the limited amount of fire tests available in the structural fire engineering domain, this paper proposed two methods to facilitate ML into structural fire engineering and other applications that utilize validated ML ensembles, Monte Carlo and Bayesian simulations to generate synthetic and augmented observations are being tested as they might prove handy in the future. Future works are to strive to develop much larger databases with higher dimensionality to enable fire designers and engineers from attaining additional insights into the workings of fire-related phenomena. Given the randomness of fire, a good practice that can significantly improve ML predictions is for future fire testing to allocate duplicated (repeatability) testing.

Future research works are invited to target developing coding-free ML methods to allow stakeholders (i.e., government agents and practitioners etc.) with limited coding experience from adopting and applying ML tools in structural fire engineering, safety and design applications. A key research need that the community needs to create standardized ML procedures to develop and validate ML algorithms and tools to further facilitate a harmonious acceptance of ML and facilitate wider adoption of this modern technology.

## 6.0 CONCLUSIONS

This paper presented the development of a novel ML ensemble to enable explainable and rapid assessment of fire resistance and fire-induced spalling of reinforced concrete (RC) columns. The developed approach comprises of an ensemble of three algorithms namely; random forest (RF), extreme gradient boosted trees (ExGBT), and deep learning (DL). The developed ML ensemble has been calibrated and validated for columns is subjected to standard and design fire exposures or under one-, two-, three- or four-sided fire exposure. The following conclusions could also be drawn from the results of this study:

- ML can provide a modern and robust tool to evaluate fire-induced phenomena in structural members. For example, the developed ML ensemble is capable of analyzing the susceptibility to fire-induced spalling and fire resistance of 350 and 85 RC columns per second, respectively.
- The ML analysis identified concrete cover thickness, compressive strength and geometric size of the column to be of highest importance to the spalling phenomenon. The same analysis also shows that concrete cover, magnitude of applied loading, and resistant conditions to have higher influence of fire resistance of columns.
- The developed ensemble can be extended to cover other structural members or fire conditions.
- Some of these challenges that may limit full integration of ML into structural fire engineering application can be overcome through two newly proposed techniques to use synthetic and augmented observations, as well as through future efforts designed to leverage ML into fire tests and simulations.